\documentclass[sigconf]{acmart}

\AtBeginDocument{%
  }

\setcopyright{acmlicensed} 
\copyrightyear{2023}
\acmYear{2023}
\acmDOI{10.1145/3581783.3613461} 

\acmConference[MM '23] {Proceedings of the 31st ACM International Conference on Multimedia}{October 29--November 3, 2023}{Ottawa, ON, Canada.}
\acmBooktitle{Proceedings of the 31st ACM International Conference on Multimedia (MM '23), October 29--November 3, 2023, Ottawa, ON, Canada}
\acmPrice{15.00}
\acmISBN{979-8-4007-0108-5/23/10}

\newcommand{\etal}{\textit{et al}.}
\newcommand{\ie}{\textit{i}.\textit{e}.}

\usepackage{multirow}
\usepackage{makecell}
\usepackage{float}
\usepackage[american]{babel}
\usepackage{microtype}
\usepackage{subfig}
\usepackage{balance}

\begin{document}

\title{Hawkeye: A PyTorch-based Library for Fine-Grained Image Recognition with Deep Learning}\thanks{This work was supported by National Key R\&D Program of China (2021YFA1001100), National Natural Science Foundation of China under Grant (62272231, 62201265), Natural Science Foundation of Jiangsu Province of China under Grant (BK20210340). Corresponding authors: X.-S. Wei and Y. Wu.}


\author{Jiabei He}
\affiliation{
  \institution{Nanjing University of Science and Technology}
  \city{Nanjing}
  \country{China}
}
\email{hejb@njust.edu.cn}

\author{Yang Shen}
\affiliation{
  \institution{Nanjing University of Science and Technology}
  \city{Nanjing}
  \country{China}
}
\email{shenyang_98@njust.edu.cn}

\author{Xiu-Shen Wei}
\affiliation{
  \institution{Southeast University}
  \city{Nanjing}
  \country{China}
}
\email{weixs.gm@gmail.com}

\author{Ye Wu}
\affiliation{
  \institution{Nanjing University of Science and Technology}
  \city{Nanjing}
  \country{China}
}
\email{wuye@njust.edu.cn}

\begin{abstract}
Fine-Grained Image Recognition (FGIR) is a fundamental and challenging task in computer vision and multimedia that plays a crucial role in Intellectual Economy and Industrial Internet applications. However, the absence of a unified open-source software library covering various paradigms in FGIR poses a significant challenge for researchers and practitioners in the field. To address this gap, we present \emph{Hawkeye}, a PyTorch-based library for FGIR with deep learning. \emph{Hawkeye} is designed with a modular architecture, emphasizing high-quality code and human-readable configuration, providing a comprehensive solution for FGIR tasks. In \emph{Hawkeye}, we have implemented 16 state-of-the-art fine-grained methods, covering 6 different paradigms, enabling users to explore various approaches for FGIR. To the best of our knowledge, \emph{Hawkeye} represents the first open-source PyTorch-based library dedicated to FGIR. It is publicly available at \url{https://github.com/Hawkeye-FineGrained/Hawkeye/}, providing researchers and practitioners with a powerful tool to advance their research and development in the field of FGIR.
\end{abstract}

\ccsdesc[500]{Software and its engineering~Software libraries and repositories}
\ccsdesc[500]{Computing methodologies~Computer vision}

\keywords{Open-Source, Fine-Grained Image Recognition, Library, Deep Learning, Convolutional Neural Networks}


\maketitle

\section{Introduction}

In recent years, significant advancements have been made in deep learning design and training, leading to substantial improvements in image recognition performance on large-scale datasets. Fine-Grained Image Recognition (FGIR) is a specialized area of research that focuses on the visual recognition of subcategories at a highly granular level within a broader semantic category. Despite significant progress with the help of deep learning~\cite{wei_survey}, FGIR remains a highly challenging task.
Also, it has significant scientific and practical applications in various scenarios within the Intellectual Economy and Industrial Internet, such as smart city, public safety, ecological protection, agricultural production and safety assurance, etc.
The main challenge in FGIR is to understand the subtle visual differences that are necessary to distinguish objects with highly similar overall appearances but differing fine-grained features. 
The primary methods of FGIR can be roughly grouped into three paradigms~\cite{wei_survey}: (1) recognition by localization-classification subnetworks, (2) recognition by end-to-end feature encoding, and (3) recognition with external information.

Despite some methods from these paradigms being open-sourced, there is currently no unified open-source library available.
New researchers in the field face a significant hindrance in replicating new approaches because different methods use distinct deep learning frameworks and design architectures, requiring the researchers to familiarize themselves with a new set of frameworks every time.
Moreover, the absence of a unified library often necessitates researchers to develop the underlying code themselves, resulting in a waste of valuable time.
Additionally, it is challenging to compare research results since each researcher/developer uses a distinct framework and base setup, leading to less reproducible results. 
Consequently, a unified open-source library is crucial for advancing the field of FGIR. 
To address this need, we developed a PyTorch-based library for FGIR, termed as \emph{Hawkeye}.


\begin{figure*}[htbp]
	\centering
	\subfloat[]{\includegraphics[width=0.98\linewidth]{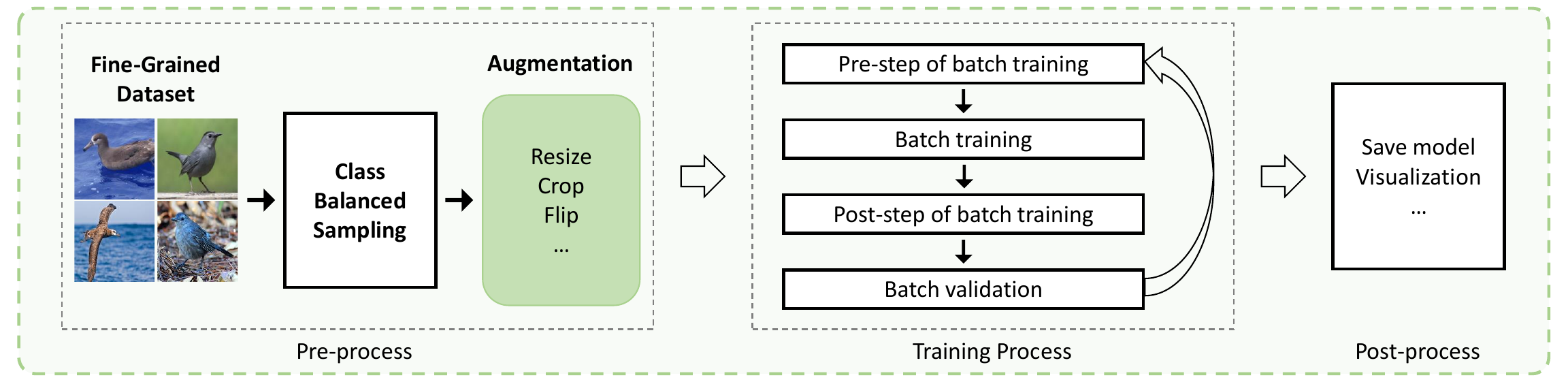}\label{fig:framework_a}}\\
	\subfloat[]{\includegraphics[width=0.98\linewidth]{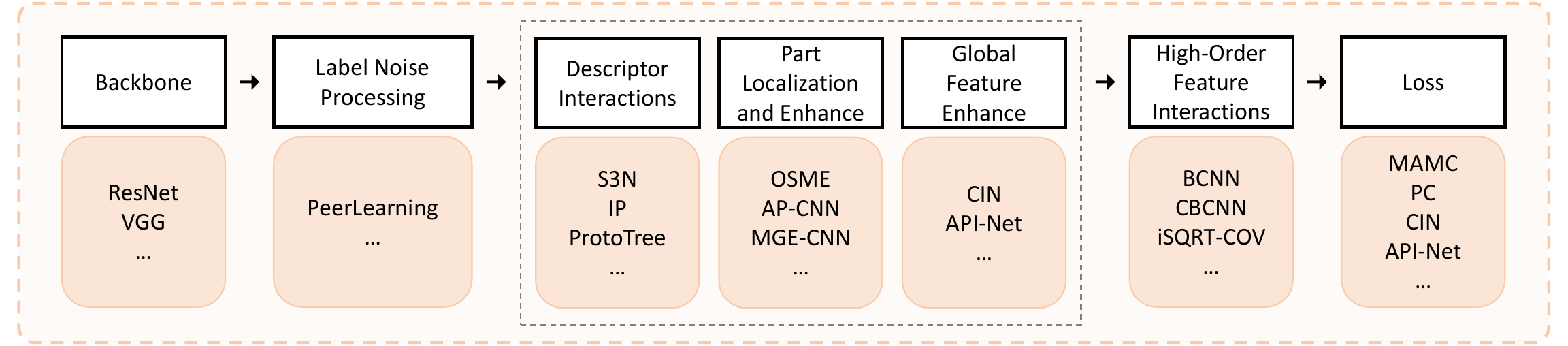}\label{fig:framework_b}}
        \vspace{-0.4em}
	\caption{(a) Workflow of \emph{Hawkeye}, including three stages: data pre-process, model training process and post-process. (b)~Modules in \emph{Hawkeye}. We present fine-grained methods involved in \emph{Hawkeye} which mainly correspond to specific modules.}
    \label{fig:framework}
\end{figure*}

The design of our work has the following advantages:

$\bullet$ \quad To the best of our knowledge, this is the first dedicated codebase designed specifically for FGIR. Our library encompasses 16 representative methods spanning 6 paradigms in FGIR, providing researchers with a comprehensive understanding of the current state-of-the-art techniques in the field. Furthermore, our modular design allows researchers to easily integrate their own methods or enhancements into the library, facilitating fair comparisons with other approaches. We are committed to maintaining and updating the open-source library to accommodate new advancements and emerging fine-grained methods in the future.

$\bullet$ \quad Modular Design: The fine-grained recognition training pipeline is structured into multiple modules, which are subsequently integrated into a unified pipeline within the \texttt{Trainer} class. Users have the flexibility to override specific modules by inheriting the \texttt{Trainer} class, allowing for customization as per their requirements. The implementation of most methods does not necessitate extensive code modifications, ensuring that \emph{Hawkeye} remains both flexible and user-friendly.

$\bullet$ \quad High Code Quality: Our library prioritizes code readability in the pipeline implementation, emphasizing the simplification of each module and ensuring the complete comprehensibility of the pipeline process. This approach enables beginners to quickly familiarize themselves with the training process of fine-grained recognition and the functions of each part.

$\bullet$ \quad Human-readable Configuration: Our library provides configuration files in the \texttt{YAML} format for each method. These files allow users to easily modify all the necessary hyperparameters for training, including those related to the dataset, model, optimizer, scheduler, and more. By focusing on a single configuration file, users can conveniently customize their experiments and adjust various settings to suit their specific needs.

\section{Design Overview}

The workflow and composition modules of \emph{Hawkeye} are illustrated in Figure~\ref{fig:framework}. The pipeline of fine-grained image recognition is grouped into several modules, shown in black blocks of Figure~\ref{fig:framework_b}. Next, we will introduce the relationship of modules to paradigms in~\cite{wei_survey}, essential modules, and the architecture of \emph{Hawkeye}.

\subsection{Correspondence between Modules and Learning Paradigms}

Basically, the modules in \emph{Hawkeye} correspond to the learning paradigms of fine-grained recognition~\cite{wei_survey}. Specifically, the label noise processing module corresponds to methods of the ``recognition with web data'' paradigm. The descriptor interactions module corresponds to methods of the ``recognition by utilizing deep filters'' paradigm. The part localization and enhancement module is mainly composed of methods from ``recognition by leveraging attention mechanisms''. The high-order feature interactions module is made up of methods from the ``recognition by performing high-order feature interactions'' paradigm. The loss function module corresponds to ``recognition by designing specific loss functions''.

\subsection{Composition Modules in \emph{Hawkeye}}

We present the composition modules in \emph{Hawkeye} as follows.

$\bullet$ \quad \textbf{Class Balanced Sampling:} It samples data in the pre-process stage. This module is essential for methods that compare different classes of samples and require balanced sampling of multiple classes of samples in a single batch.

$\bullet$ \quad \textbf{Backbone:} This module provides basic feature extraction networks, including ResNet and VGG.

$\bullet$ \quad \textbf{Label Noise Processing:} It focuses on the process of handling label noise in webly fine-grained images, leaving clean data for subsequent modules.
    
$\bullet$ \quad \textbf{Descriptor Interactions:} It leverages the locality and spatiality of descriptors to detect parts of fine-grained objects.

$\bullet$ \quad \textbf{Part Localization and Enhancement:} This module detects the parts of a fine-grained object and constructs part-level representations corresponding to those parts, considering the small differences among fine-grained categories.

$\bullet$ \quad \textbf{Global Feature Enhancement:} It explores interactions between deep channels or pairs of images using image-level representations.

$\bullet$ \quad \textbf{High-Order Feature Interactions:} This module encodes the second-order statistics derived from convolutional activations.

$\bullet$ \quad \textbf{Loss:} This module directly drives classifier learning and image representation learning through a loss function designed for fine-grained recognition.

\subsection{Architecture}



Each method has a configuration file in the \texttt{YAML} format that can be easily modified for specific parameters. 
The \texttt{Trainer} class implements the core functions for training, such as batch training methods, optimizers, hooks, and checkpoints.
Users can implement their customized methods by inheriting the \texttt{Trainer} class, and a few lines of code require to be modified. 
A generic \texttt{Dataset} class is implemented for different fine-grained datasets.
With the meta-data files provided in \emph{Hawkeye}, users can easily apply and switch between the eight fine-grained benchmark datasets in experiments. 
The \texttt{Model} module includes the specific implementation of various methods, as well as the special \texttt{Loss} required by some methods. 
Users can easily add their own methods to \emph{Hawkeye}. 
These modules are designed to be expandable, allowing users to implement customized designs without modifying unnecessary code.

\section{Supported Methods}

In this paper, we provide the following representative fine-grained recognition methods of 6 different types according to \cite{wei_survey}, including utilizing deep filters, leveraging attention mechanisms, performing high-order feature interactions, designing specific loss functions, recognizing with web data, as well as miscellaneous.
We have chosen 16 representative methods from these 6 types and implemented them in the library. We will briefly introduce these methods.

$\bullet$ \quad S3N~\cite{s3n} leverages class peak responses, \ie, local maximums, as the basis of part localization, based on class response maps~\cite{cam}.

$\bullet$ \quad IP~\cite{ip} provides an interpretation of classification results via the segmentation of object parts and the identification of their contributions.

$\bullet$ \quad ProtoTree~\cite{prototree} combines prototype learning with decision trees, and thus results in an intrinsically interpretable model.

$\bullet$ \quad MGE-CNN~\cite{mgecnn} promotes diversity among a mixture of experts by combing an expert gradually-enhanced learning strategy and a Kullback-Leibler divergence-based constraint.

$\bullet$ \quad Sun~\etal~\cite{mamc} incorporates channel attentions and metric learning to enforce the correlations among attended regions.
    
$\bullet$ \quad APCNN~\cite{apcnn} integrates low-level information to obtain enhanced feature representation and accurately located discriminative regions using a pyramidal hierarchy structure.

$\bullet$ \quad Bilinear~CNN~\cite{bcnn} leverages bilinear pooling over the outputs of two CNNs to model local pairwise feature interactions in a translationally invariant manner.

$\bullet$ \quad CBCNN~\cite{cbcnn} utilizes two compact bilinear representations with the same discriminative power as the full bilinear representation but with only a few thousand dimensions.

$\bullet$ \quad Fast MPN-COV~\cite{fastmpn} (\ie, iSQRT-COV) proposes an iterative matrix square root normalization method for fast end-to-end training of global covariance pooling networks.

$\bullet$ \quad PC~\cite{pc} reduces overfitting by intentionally introducing confusion in the activations.

$\bullet$ \quad API-Net~\cite{apinet} attentively captures contrastive clues by pairwise interaction between two images.

$\bullet$ \quad CIN~\cite{cin} models the channel-wise interplay within and across images to exploit the rich relationships between channels.

$\bullet$ \quad PeerLearning~\cite{peerlearning_webfg} trains two deep neural networks simultaneously, both of which mutually communicate proper knowledge from noisy web images. 

$\bullet$ \quad NTSNet~\cite{ntsnet} localizes informative regions with Navigator, Teacher and Scrutinizer cooperating and reinforcing each other.

$\bullet$ \quad Cross-X~\cite{crossx}  exploits the relationships between different images and between different network layers for robust multi-scale feature learning.

$\bullet$ \quad DCL~\cite{dcl} destructs and then reconstructs the fine-grained image, for learning discriminative regions and features.

\section{Experiments}


\subsection{Benchmark Datasets}

Eight representative fine-grained recognition benchmark datasets are provided. We provide the meta-data file of the datasets, and the \textit{train} list and the \textit{val} list are also provided according to the official splittings of the dataset. 
Researchers can easily utilize these datasets by following the examples provided in the library. 
Table~\ref{tab:datasets} provides a summary of the year of publication, meta-category, number of images, and number ofcategories for each dataset.




\subsection{Implementation Details}
In our implementation, we use a NVIDIA GeForce 3060 GPU to train and infer the models of each method.
We perform the training stage mainly using the image size of $448\times448$. 
The batch size, learning rate and epoch were set according to each method, using as many settings as possible from the method's corresponding paper, and these settings are detailed in individual config files for each method.
We initializes the models with ResNet and VGG weights pre-trained on ImageNet, except for ProtoTree~\cite{prototree}, which uses weights pre-trained on iNat2017~\cite{inat2017}. 
The optimisers are mainly Stochastic Gradient Descent (SGD), Adaptive Moment Estimation (Adam) or Adam with decoupled weight decay (AdamW).
Most methods are trained using cosine learning rate scheduler with warm-up function to adjust the learning rate, while others use  step learning rate or multiple step learning rate, etc. 
Image augmentation methods include random resized crop, random horizontal flip and random erasing with a probability of 0.1.

\begin{table}[]
\setlength{\tabcolsep}{2pt}
\caption{Fine-grained benchmarks provided in \emph{Hawkeye}.}
\vspace{-1.2em}
\label{tab:datasets}
\begin{tabular}{l|c|c|c|c}
\hline
\textbf{Dataset} & \textbf{Year} & \textbf{Meta-class} & \textbf{\# images} & \textbf{\# categories} \\ \hline
CUB-200~\cite{cub200} & 2011 & Birds & 11,788 & 200 \\
Stanford Dog~\cite{stanford_dogs} & 2011 & Dogs & 20,580 & 120 \\
Stanford Car~\cite{stanford_cars} & 2013 & Cars & 16,185 & 196 \\
FGVC Aircraft~\cite{aircraft} & 2013 & Aircrafts & 10,000 & 100 \\
iNat2018~\cite{inat2017} & 2018 & \makecell{Plants \&\\ Animals} & 461,939 & 8,142 \\
WebFG-bird~\cite{peerlearning_webfg} & 2021 & Birds & 18,388 & 200 \\
WebFG-car~\cite{peerlearning_webfg} & 2021 & Cars & 21,448 & 196 \\
WebFG-aircraft~\cite{peerlearning_webfg} & 2021 & Aircrafts & 13,503 & 100 \\ \hline
\end{tabular}
\end{table}

\begin{table}[]
\caption{Performence of fine-grained recognition methods on the CUB-200 dataset. Except for the asterisked methods, $ 448\times448 $ input images were used. }
\vspace{-1em}
\label{tab:results}
\begin{tabular}{lcc}
\hline
\textbf{Methods} & \textbf{Original Acc.} & \textbf{Acc. in Hawkeye} \\ \hline
\multicolumn{3}{l}{\textit{\textbf{Utilizing Deep Filters}}} \\ \hline
S3N~\cite{s3n} & 88.50 & 88.29 \\ 
IP~\cite{ip} & 87.30 & 86.65 \\
ProtoTree~\cite{prototree} & 82.20* & 82.94* \\ \hline
\multicolumn{3}{l}{\textit{\textbf{Leveraging Attention Mechanisms}}} \\ \hline
MGE-CNN~\cite{mgecnn} & 88.50 & 89.05 \\
OSME+MAMC~\cite{mamc} & 86.50 & 84.31* \\
APCNN~\cite{apcnn} & 88.40 & 87.84 \\ \hline
\multicolumn{3}{l}{\textit{\textbf{Performing High-Order Feature Interactions}}} \\ \hline
BCNN~\cite{bcnn} & 84.10 & 83.80 \\
CBCNN~\cite{cbcnn} & 84.00 & 84.13 \\
Fast MPN-COV~\cite{fastmpn} & 88.10 & 88.81 \\ \hline
\multicolumn{3}{l}{\textit{\textbf{Designing Specific Loss Functions}}} \\ \hline
Pairwise Confusion~\cite{pc} & 80.21 & 87.67 \\
API-Net~\cite{apinet} & 87.70 & 87.88 \\
CIN~\cite{cin} & 87.50 & 85.34* \\ \hline
\multicolumn{3}{l}{\textit{\textbf{Recognition with Web Data}}} \\ \hline
Peer-Learning~\cite{peerlearning_webfg} & 76.48 & 77.85 \\ \hline
\multicolumn{3}{l}{\textit{\textbf{Miscellaneous}}} \\ \hline
NTS-Net~\cite{ntsnet} & 87.50 & 88.19 \\
CrossX~\cite{crossx} & 87.70 & 87.65 \\
DCL~\cite{dcl} & 87.80 & 87.64 \\ \hline
\end{tabular}
\vspace{-1em}
\end{table}

\subsection{Results}

We have conducted experiments on the methods implemented in \emph{Hawkeye} using CUB-200~\cite{cub200} to prove the effectiveness of our library. 
By integrating these methods with different implementations into a unified fine-grained recognition framework, some results show slight fluctuations, but they are still within acceptable limits.

We categorized the results based on the paradigm in \cite{wei_survey} for easy observation and analysis, as presented in Table~\ref{tab:results}. 
Most of our experiments are performed on $ 448\times448 $ input images, and the results marked with an asterisk use $224\times224$ input images.

\vspace{-0.5em}

\section{Availability}
\emph{Hawkeye} is released under the license of MIT and available at: \href{https://github.com/Hawkeye-FineGrained/Hawkeye/}{https://github.com/Hawkeye-FineGrained/Hawkeye/}.
We also provide documentation and training samples. Contributions from the open-source community are welcome, via the GitHub issues/pull request mechanism.

\section{Conclusions}

We developed \emph{Hawkeye}, the first open-source PyTorch-based library for fine-grained recognition with deep learning. Featuring a modular design, our library ensures simplicity and ease of extension. Each method is accompanied by training examples that require only minor code modifications, showcasing the user-friendly and highly adaptable nature of \emph{Hawkeye}. We have implemented 16 fine-grained methods in a unified framework. It facilitates researchers in rapidly acquainting themselves with the cutting-edge advancements in fine-grained recognition, and expediting their exploration of novel ideas and enhancements We are dedicated to the ongoing maintenance and refinement of \emph{Hawkeye} as an open-source project.



\bibliographystyle{ACM-Reference-Format}
\balance
\bibliography{main}

\end{document}